\documentclass[11pt]{article}

\usepackage{acl}
\usepackage{times}
\usepackage{latexsym}
\usepackage[T1]{fontenc}
\usepackage[utf8]{inputenc}
\usepackage{microtype}
\usepackage{inconsolata}
\usepackage{adjustbox}
\usepackage{fancyhdr}
\usepackage{amsmath,amssymb,amsfonts}
\usepackage{booktabs}
\usepackage{multirow}
\usepackage{threeparttable}
\usepackage{graphicx}
\usepackage{subcaption}
\usepackage{enumitem}
\usepackage{algorithm}
\usepackage{algpseudocode}
\graphicspath{{figure/}{figures/}}

\IfFileExists{math_commands.tex}{

\usepackage{amsmath,amsfonts,bm}

\def\secref#1{section~\ref{#1}}

\def\eqref#1{equation~\ref{#1}}

\def\1{\bm{1}}

\DeclareMathAlphabet{\mathsfit}{\encodingdefault}{\sfdefault}{m}{sl}
\SetMathAlphabet{\mathsfit}{bold}{\encodingdefault}{\sfdefault}{bx}{n}

}{}

\newcommand{\appref}[1]{Appendix~\ref{#1}}

\newcommand{\best}[1]{\textbf{#1}}
\newcommand{\secondbest}[1]{\underline{#1}}
\title{The Hidden Costs and Measurement Gaps of Reinforcement Learning with Verifiable Rewards}

\author{
{\bfseries
Fang Wu\textsuperscript{1*\textdagger} \quad
Aaron Tu\textsuperscript{2*} \quad
Weihao Xuan\textsuperscript{3,4*} \quad
Heli Qi\textsuperscript{4,5*} \quad
Xu Huang\textsuperscript{6} \quad
Qingcheng Zeng\textsuperscript{7}} \\
{\bfseries
Shayan Talaei\textsuperscript{1} \quad
Yijia Xiao\textsuperscript{8} \quad
Peng Xia\textsuperscript{9} \quad
Xiangru Tang\textsuperscript{10} \quad
Yuchen Zhuang\textsuperscript{6} \quad
Yinxi Li\textsuperscript{11}} \\
{\bfseries
Bing Hu\textsuperscript{12} \quad
Hanqun Cao\textsuperscript{13} \quad
Wenqi Shi\textsuperscript{14} \quad
Rui Yang\textsuperscript{15} \quad
Nan Liu\textsuperscript{15} \quad
Huaxiu Yao\textsuperscript{9}} \\
{\bfseries
Ge Liu\textsuperscript{16} \quad 
Li Erran Li\textsuperscript{17} \quad
Amin Saberi\textsuperscript{1} \quad
Naoto Yokoya\textsuperscript{3,4} \quad
Jure Leskovec\textsuperscript{1} \quad
Yejin Choi\textsuperscript{1\textdagger}} \\[2pt]
\textsuperscript{1}Stanford University \quad
\textsuperscript{2}UC Berkeley \quad
\textsuperscript{3}The University of Tokyo \quad
\textsuperscript{4}RIKEN AIP \\
\textsuperscript{5}Waseda University \quad
\textsuperscript{6}Georgia Tech \quad
\textsuperscript{7}Northwestern University \quad
\textsuperscript{8}UCLA \\
\textsuperscript{9}UNC Chapel Hill \quad
\textsuperscript{10}Yale University \quad
\textsuperscript{11}University of Waterloo \\
\textsuperscript{12}Independent Researcher \quad 
\textsuperscript{13}CUHK \quad
\textsuperscript{14}UT Southwestern Medical Center \\
\textsuperscript{15}National University of Singapore  \quad 
\textsuperscript{16}UIUC \quad
\textsuperscript{17}Amazon AWS AI \\[2pt]
\textsuperscript{*}Equal Contribution \qquad
\textsuperscript{\textdagger}Corresponding Authors
}

\begin{document}
\maketitle
\begin{abstract}
Reinforcement learning with verifiable rewards (RLVR) is a practical, scalable way to improve large language models on math, code, and other structured tasks. However, we argue that many headline RLVR gains are not yet well validated because reports often conflate policy improvement with three confounds: (i) budget mismatch between RLVR and baseline evaluations, (ii) attempt inflation and calibration drift that convert abstentions into confident answers, and (iii) benchmark data contamination. Using budget-matched reproductions and partial-prompt contamination probes, we find that several widely cited gaps shrink substantially or disappear once budgets, prompts, and dataset versions are matched and contaminated sets are treated as memorization probes rather than evidence of reasoning. This does not mean that RLVR is ineffective, but it implies that current measurements often overstate capability gains and obscure reliability costs. We therefore propose a compact, tax-aware minimum standard for RLVR training and evaluation: budget-matched saturation curves with variance, calibration, and abstention tracking, a judge-robustness stress test when LLM judges are used, and an explicit contamination screen. With these controls, RLVR remains effective and deployable in verifiable domains, but reasoning gains should be treated as provisional without them.
\end{abstract}


\section{Introduction}
\label{sec:intro}

Reinforcement learning with verifiable rewards (RLVR) has become a leading post-training route for improving large language models (LLMs) on math and code tasks~\citep{luong2024reft, wen2025light}. By optimizing against automatically computable signals such as unit tests for programs, exact numeric or string matches for math, or retrieval-grounded checks for citations, RLVR promises a path to better performance. Recent work reports large benchmark gains across domains. Appendix Fig.~\ref{fig:rlvr_trend} shows a trend plot of RLVR-tagged paper activity and AIME24/25 performance with striking monthly gains. Our position is that RLVR is effective and deployable for verifiable domains, but ``reasoning'' gains should be treated as provisional without a small core of tax-aware controls.

A central question frames the debate: \emph{Does RLVR genuinely impart new reasoning capability, or does it mainly sharpen selection among behaviors the base model already knows how to produce?} Budget-parity controlled studies show that base models can narrow or erase RLVR gaps when given matched sampling budgets, consistent with improved search rather than capability expansion~\citep{yue2025does,wu2025invisible}. Meanwhile, several settings report gains that are hard to recover by sampling alone: explicitly optimizing the multi-sample objective (\textit{pass@k} training)~\citep{chen2025passk}, curriculum via self-play with variational problem synthesis~\citep{liang2025svs}, and distribution-aware reward shaping that counters rank bias and diversity collapse, alongside longer-horizon RL schedules and unlikeliness rewards~\citep{liu2025prorl,he2025rewarding}. Metric choice can flip conclusions: answer-only scores may diverge from process-aware metrics that require both a correct answer and a valid chain of thought.

A second complication is what we call the \emph{RLVR tax}: unintended \emph{empirical} side effects that accompany apparent gains under current reasoning-style post-training (reduced abstention, miscalibration, instruction-fidelity drift, and a larger safety/privacy surface due to longer traces). This tax is neither inevitable nor unique to RLVR; similar patterns arise under reasoning-heavy SFT and RLHF~\citep{wu2025multiplayer}. We focus on RLVR because verifiable objectives and open-weight models make these trade-offs measurable at relatively low cost. In practice, RLVR tends to reduce abstention and increase stated confidence, sometimes even when answers are wrong, thereby shifting risk from ``I don't know'' to assertive errors~\citep{song2025hallucination,mei2025reasoning,yao2025reasoning}. It can also erode instruction fidelity on longer generations, where adhering to formats or constraints becomes harder as chains grow~\citep{fu2025scaling,li2025thinking}. Finally, longer, more explicit reasoning traces expand the attack and leakage surface, raising jailbreak success and privacy exposure if left unchecked~\citep{zhou2025hidden,jiang2025safechain,ackerman2025manyshot,green2025leakythoughtslargereasoning,huang2025safety,zhang2025should}.

\begin{figure*}[ht]
  \centering
  \includegraphics[width=\textwidth]{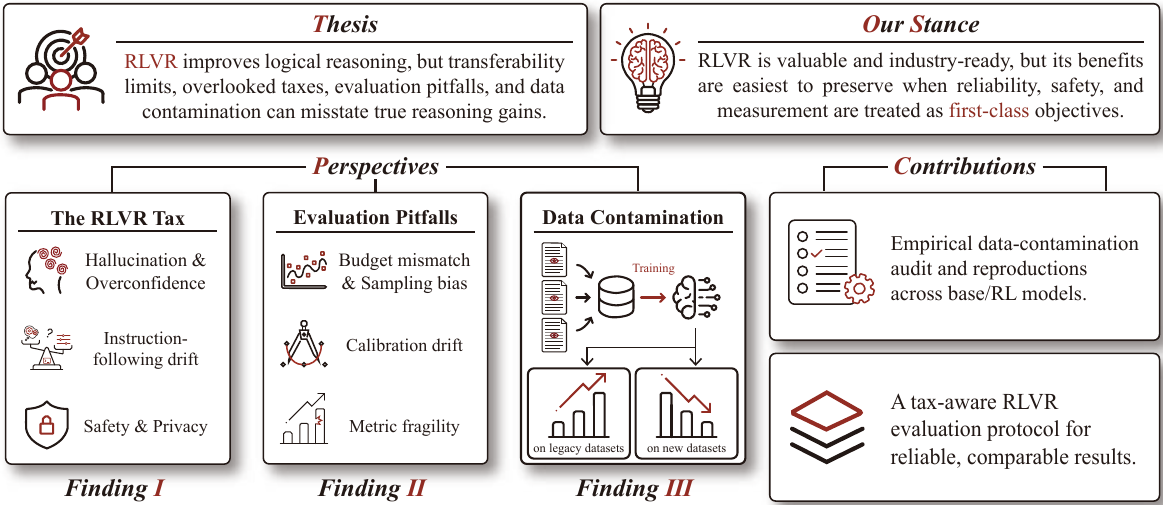}
  \caption{Paper Roadmap: taxes, evaluation pitfalls, contamination, and the unified protocol.}
  \vspace{-1em}
  \label{fig:fig1-wide}
\end{figure*}

A third complication is \emph{measurement}. Reported advances are sensitive to sampling budgets and decoding settings~\citep{hochlehnert2025sober,llmrl2025incorrect,brown2024large,muennighoff2025s1}, to the stability of LLM-as-a-judge evaluators~\citep{zhao2025onetoken,tan2024judgebench,gu2024llmasajudgesurvey}, to calibration drift~\citep{leng2024taming,xiao2025restoring}, and to data provenance~\citep{he2025deepmath,mirzadeh2024gsm,liu2024livecodebench,xu2025retrieval,wu2024insertgnn,wu2025large,wudeepsearch}. When budgets are matched, judges are stress-tested, and datasets are versioned and decontaminated, reported improvement gaps can diminish, suggesting that part of the apparent progress reflects evaluation design rather than durable capability~\citep{hochlehnert2025sober,llmrl2025incorrect}. We organize the paper around three threads: the RLVR tax, evaluation pitfalls, and data contamination, culminating in a compact tax-aware minimum standard. A visual roadmap is in Fig.~\ref{fig:fig1-wide}.

\paragraph{Scope and definitions.}
We use \emph{RLVR} to mean post-training that optimizes LLMs against \emph{automatically checkable} signals rather than human preference models~\citep{ouyang2022training}. Typical implementations adopt PPO/GRPO/DAPO-style updates with a KL penalty to the base policy, often mixing offline rollouts with online sampling, adding entropy regularization, and using selective filtering or gating. Rewards are frequently \emph{componentized} (for correctness, grounding/citation sufficiency, and calibrated refusal) and introduced in stages. Our analysis centers on open-weight models fine-tuned for math, code, and question answering (QA) with verifiable objectives, in both single- and multi-domain regimes. We deliberately avoid a catalogue of low-level algorithmic variants except where they bear on reliability (e.g., unlikeliness rewards that counter rank bias, or curricula that change variance). The primary threats considered are privacy leakage from long chains of thought (CoT) and increased jailbreak susceptibility during or after RLVR.

\paragraph{Contributions and Roadmap.}
Our contributions are threefold: \emph{(1) Measurement:} we provide budget-matched reproductions and a unified gap table that isolates the effects of budget mismatch, judge drift, template drift, and dataset versioning (Table~\ref{tab:gap_analysis}). \emph{(2) Tax:} we quantify attempt inflation and the resulting calibration and refusal costs under RLVR and matched SFT controls (Table~\ref{tab:factuality-compact}). \emph{(3) Protocol:} based on these findings, we distill a compact \emph{tax-aware minimum standard} for RLVR training and evaluation, intended as a minimum credible reporting bar, which we instantiate in \secref{sec:protocol}. The paper is organized accordingly: we (i) synthesize evidence on sharpening vs.\ expansion (\secref{sec:unlock}); (ii) analyze the RLVR tax and how it distorts reported improvements (\secref{sec:tax}); (iii) reproduce representative gaps under parity controls (\secref{sec:evaluation}); (iv) audit contamination with partial-prompt probes (\secref{sec:contamination}); and (v) present the tax-aware minimum standard (\secref{sec:protocol}).

\section{What Changes Under RLVR? An Evidence Overview}
\label{sec:unlock}

We summarize evidence under two lenses: sharpening vs broadening. Skeptical and optimistic perspectives recur in the literature, and each tends to appear under identifiable conditions.

\subsection{Skeptical Lens: Sharpening and Sampling Effects}

RLVR appears to improve sample efficiency and steer the model toward high-reward regions already present in the base distribution, rather than broaden fundamental capability~\citep{gandhi2025cognitive,shah2025rethinking}. When base models are evaluated under matched sampling budgets (i.e., \textit{pass@k}), gaps between base and RLVR-trained models often shrink~\citep{yue2025does}, consistent with sharpening rather than expansion. Empirically, outputs can collapse toward a dominant pretraining mode with reduced diversity~\citep{zhao2025echo}, and theory suggests that after RLVR, support \emph{shrinkage} tends to outweigh support \emph{expansion}~\citep{wu2025invisible}. Stress tests add caution: performance collapses at higher complexities have been reported even with ample compute~\citep{shojaee2025illusion}, although some failures trace to evaluation artifacts such as token limits and unsolvable items, underscoring the need for budget parity, seed control, and dataset hygiene (see \secref{sec:evaluation}). Finally, several works separate \emph{knowledge injection} from \emph{policy optimization}: pipelines that use SFT or distillation to inject knowledge and RLVR to select among candidates typically beat pure GRPO, suggesting RLVR optimizes within learned support~\citep{ma2025learning,wan2025srpo,chen2025bridging}; guidance and self-distillation further help over vanilla GRPO~\citep{nath2025adaptiveguidanceacceleratesreinforcement,liu2025superrl}.

\subsection{Optimistic Lens: Signals of Capability Broadening}

Carefully designed RLVR appears to \emph{expand} what models can do, not merely sharpen selection. Prolonged optimization with explicit KL control and periodic resets (ProRL) reports trajectories that are not reachable by sampling the base policy alone, with the largest gains where the base initially struggles~\citep{liu2025prorl}. Reward shaping also matters: by diagnosing a GRPO tendency to reinforce high-probability completions, \citet{he2025rewarding} shows that adding unlikeliness rewards and training deeper widens the model's support and improves multi-sample theorem proving, indicating qualitative changes to reachable reasoning modes. These effects are conditioned by pretraining coverage: cross-domain analyses find broader benefits in Math/Code/Science, while under-represented areas such as Logic/Simulation/Tabular realize meaningful gains when RLVR is applied in-domain~\citep{cheng2025revisiting}. Consistent with this, process-aware metrics that require both a correct answer and a syntactically valid chain (CoT-\textit{pass@k}) often reveal larger RLVR deltas than answer-only measures~\citep{wen2025reinforcementlearningverifiablerewards}. While generalization to substantially harder regimes remains an open goal~\citep{sun2025omegallmsreasonoutside}, the weight of evidence supports a constructive view: with sufficient optimization depth, bias-aware rewards, and the right domain coverage, RLVR can broaden reasoning capabilities.

\paragraph{Reporting note.}
Answer-only metrics (e.g., \textit{pass@k}) can show muted gains, whereas process-aware criteria such as CoT-\textit{pass@k} (which require both a correct answer and a syntactically valid chain) often reveal larger deltas in settings explicitly optimized for process rewards~\citep{wen2025reinforcementlearningverifiablerewards}. We treat these as \emph{behavioral} signals rather than explanations: chains need not be faithful to internal computation~\citep{chen2025reasoning}. Accordingly, we report both answer- and process-aware metrics alongside calibration (ECE/abstention), and we prefer budget-matched saturation curves over single-point \textit{pass@k} (\secref{sec:evaluation}).

\paragraph{Cross-Lens Lessons:}
Across both perspectives, two themes are consistent: \emph{(i)} the base model's strength and coverage strongly condition observed gains, and \emph{(ii)} reward/metric design and evaluation protocol (sampling budgets, seeds, and contamination hygiene) can change the sign of conclusions. These observations motivate our position.

\section{Perspective I: The RLVR Tax: Benefits, Costs, and Controls}
\label{sec:tax}

RLVR reliably lifts accuracy on verifiable tasks, but those gains are easy to misread if we ignore systematic costs---the \emph{RLVR tax}. We elaborate on three recurrent pressures: \emph{(i)} hallucination and overconfidence, \emph{(ii)} erosion of instruction following, and \emph{(iii)} safety and privacy exposure. Each maps directly to our position: without tax-aware controls and sober measurement, reported ``reasoning gains'' are easy to overstate and hard to transfer.

\begin{table*}[t]
\centering
\caption{Factual-QA control tracking Abstention ($\downarrow$), Shared Accuracy ($\uparrow$), and ECE ($\downarrow$).
\emph{Interpretation.} ``Not attempted'' is the \emph{absolute} number of items with no extractable answer, computed on the
block-specific evaluation set for that family (totals differ across families). At 14B/32B, reasoning SFT sharply
reduces abstentions but leaves shared accuracy flat and worsens calibration; at a larger scale (DeepSeek), RL
reduces abstention and improves shared accuracy and ECE, largely by \emph{attempting more items}. Accuracy on
\emph{newly attempted} tail items is modest.}
\label{tab:factuality-compact}
\setlength{\tabcolsep}{6pt}
\renewcommand{\arraystretch}{1.15}
\resizebox{\textwidth}{!}{%
\begin{tabular}{l|l| ccc}
\toprule
\textbf{Family/Scale} & \textbf{Model (version)} &
\textbf{Not attempted $\downarrow$} &
\textbf{Accuracy (shared, \%) $\uparrow$} &
\textbf{ECE (shared) $\downarrow$} \\
\midrule
\multirow{3}{*}{Qwen2.5\,14B}
  & Qwen2.5-14B-Instruct        & 1136 & 12.5 & 0.598 \\
  & R1-Distill-Qwen-14B (SFT)   &  102 & 10.5 & 0.692 \\
  & RL-Reasoning-14B            &  103 & 10.0 & 0.684 \\
\midrule
\multirow{3}{*}{Qwen2.5\,32B}
  & Qwen2.5-32B-Instruct        & 2492 & 17.4 & 0.591 \\
  & R1-Distill-Qwen-32B (SFT)   &   76 & 17.5 & 0.640 \\
  & RL-Reasoning-32B            &   63 & 17.1 & 0.600 \\
\midrule
\multirow{2}{*}{DeepSeek}
  & DeepSeek-V3 (Instruct)      &  480 & 27.5 & 0.496 \\
  & DeepSeek-R1 (RL)            &   81 & 34.6 & 0.317 \\
\bottomrule
\end{tabular}}
\end{table*}

\subsection{Hallucination and Overconfidence}
RLVR can suppress refusals while amplifying confident errors. Empirically, refusal rates often collapse after RLVR, shifting abstentions into assertive answers~\citep{song2025hallucination}; models may also repeat flawed steps or produce CoT that diverges from the final answer even when accuracy rises~\citep{yao2025reasoning}. One hypothesis is that sparse, verifiable rewards combined with entropy pressure tend to encourage determinacy under weak evidence, producing high-variance gradients and spurious local optima~\citep{li2025hallucination}. Consistent with this, multiple studies find substantial miscalibration: self-reported confidence remains high even on \emph{incorrect} responses and can increase with longer CoT traces~\citep{mei2025reasoning,zeng2025thinking,kirichenko2025abstentionbench}.

We make this concrete with a factual-QA control that tracks \emph{Not attempted} (lower is better), \emph{Accuracy (shared)} (accuracy on items that both models at a given scale attempted; higher is better), and \emph{Expected Calibration Error (ECE; lower is better)} computed on shared items from the model's stated confidence. The dominant effect is \emph{attempt inflation}: stronger RLVR policies attempt far more questions. For example, moving from \textbf{DeepSeek-V3} to \textbf{DeepSeek-R1}~\citep{guo2025deepseek} sharply reduces abstentions and improves shared ECE, but accuracy on the \emph{newly attempted} tail is modest, so reported gains depend on how one aggregates across attempted and unattempted items~\citep{zeng2025thinking}. In other words, scaling RL \emph{encourages attempts}; whether this yields net quality improvements depends on calibration and gating.

\subsection{Instruction Following}
Reasoning-optimized training can erode controllability, especially for long generations. Several studies observe regressions in instruction fidelity when training emphasizes extended chain-of-thought or purely verifiable endpoints~\citep{fu2025scaling,li2025thinking}. At the same time, recent evidence shows that models tend to overfit to a small set of verifiable constraints and struggle to generalize; IFBench (58 out-of-domain constraints) documents this gap and also finds gains under RLVR in precise instruction-following generalization~\citep{pyatkin2025ifbench}. This is why our minimum standard in \secref{sec:protocol} requires either an explicit instruction or format component in the reward, or at least a separate instruction-following evaluation pack.

\subsection{Safety and Privacy}
Long, explicit reasoning traces widen the attack and leakage surface. Frontier reasoning models are jailbreakable at high rates; automated attacks approach deterministic success in the studied settings, and success rises with attempt and context budgets~\citep{KassianikKarbasi2025,zhou2025hidden,jiang2025safechain}. Many-shot evaluations show the same scaling. Targeted defenses tuned specifically to that regime can sharply reduce measured attack success in controlled tests, but these gains are brittle across models and budgets and can impose utility costs~\citep{ackerman2025manyshot}. Richer CoT also increases privacy risk: attributes, confidential prompts, and dataset contents can be reconstructed more easily as reasoning traces grow longer~\citep{green2025leakythoughtslargereasoning}. Safety tuning itself can impose a ``safety tax,'' degrading math and coding unless staged and balanced carefully~\citep{huang2025safety,zhang2025should}.

\begin{quote}\small
\noindent\textbf{Threat Model (Reasoning Traces).} \emph{Who sees traces:} exposure risk is highest for open-weight models or deployments that return full CoT; API-only hidden traces lower user-facing risk, while internal logs can aid auditing.
\emph{Training vs.\ inference:} RLVR increases trace length and attempt budgets during training and (unless capped) inference, which can amplify leakage/jailbreak surfaces.
\emph{What ``risk increase'' means:} higher probability that sensitive prompt/data fragments or unsafe behaviors appear in controllable intermediate text.
\emph{When traces help:} hidden/internal traces improve monitorability and post-hoc safety audits, so the goal is controlled exposure, not blanket removal.
\end{quote}

\paragraph{Takeaway:}
These taxes distort reported gains predictably: overconfident hallucinations inflate apparent utility, instruction drift undermines deployability, and expanded attack surfaces raise real-world risk. The answer is not to abandon RLVR but to \emph{co-optimize} for reliability: compose rewards so correctness, grounding, and calibrated abstention can all improve; manage variance and difficulty; and make calibration and provenance part of evaluation. We follow this recipe in the next sections to show where conclusions change once the tax is accounted for in \secref{sec:evaluation} and \secref{sec:protocol}.

\section{Perspective II: Pitfalls in Evaluation: Are We Measuring Real Reasoning Gains Accurately?}
\label{sec:evaluation}

Claims about RLVR progress are highly sensitive to sampling budgets, metric design, and dataset hygiene. Budget mismatch, fragile judge pipelines, and calibration drift can overstate ``reasoning gains'' and complicate transfer beyond math and code. We treat contamination separately in \secref{sec:contamination} and focus here on budgeting, metric robustness, and calibration.

\subsection{Budget Parity and Saturation}
A recurring pattern in the literature is to report \textit{pass@k} for RLVR models while holding base models to much smaller budgets (e.g., \textit{pass@1} or \textit{pass@5}). In such settings, measured improvements often reflect extra search rather than a better policy. Prior work stresses matched budgets and decontaminated baselines for fair comparison~\citep{hochlehnert2025sober,llmrl2025incorrect,yue2025does}. Small or underspecified benchmarks inflate estimator variance and reduce statistical power, making single-run point estimates unstable; averaging across multiple randomized trials materially improves estimate reliability~\citep{mu2025dissecting}. Under parity controls (e.g., \emph{SoberScore}, a standardized, matched-budget, multi-seed evaluation that fixes decoding/setup and reports mean $\pm$ std), several prominent gaps shrink or disappear~\citep{hochlehnert2025sober}. In practice, evaluations should match $k$ across base and RLVR, plot saturation curves (accuracy vs.\ $k$, optionally summarized by area under the curve), and disclose decoding budgets and parameters.

\subsection{Metric Fragility and LLM Judges}
While RLVR often targets verifiable domains (math/coding), many evaluation targets either lack programmatic verifiers or have only partial/weak ones (e.g., safety/refusal appropriateness, long-form coherence and rationale quality, multi-turn task success, or fuzzy information extraction). The evaluation often relies on LLM judges. LLM-judge pipelines offer convenience but can be brittle: small changes (seed choice, $k$, dataset versioning, instruction placement, option order, or even tensor-parallel settings) produce swings comparable to reported gains~\citep{sun2025evaluation}. Judges are also manipulable~\citep{zhao2025onetoken}. Beyond judges, seemingly modest metric or setup shifts can alter conclusions about ``stable reasoning''~\citep{liu2024your}. Where programmatic verifiers exist, they should be preferred. When judges are unavoidable, prompt/format perturbations and adversarial probes should be part of the protocol, with robustness deltas reported, inter-judge agreement documented, and all templates/configurations released to reduce hidden degrees of freedom.

\subsection{Calibration Drift and Overconfidence}
RL optimization often sharpens the output distribution. Top-1 accuracy can rise even as calibration worsens, increasing brittleness under distribution shift~\citep{hochlehnert2025sober}. Combined with the volatility noted above~\citep{sun2025evaluation}, some apparent ``wins'' likely reflect confident exploitation of evaluation quirks rather than robust reasoning. Evaluations should therefore track expected calibration error (ECE), output entropy, and refusal/abstention alongside accuracy, and consider early stopping or annealing when calibration degrades even if reward continues to increase. Unless otherwise stated, numeric confidence is obtained from \emph{logit-derived} probabilities under the
model, computed from the normalized likelihood of the extracted final answer under the same decoding setup.

\begin{table*}[t]
\centering
\caption{Reported scores vs.\ standardized evaluation (\textit{avg@32} of \textit{pass@1}, estimated by averaging 32 independent single-sample decodes) with matched decoding budgets and a shared verifier/prompt family. $\Delta$ denotes Reported $-$ Standardized. See Table~\ref{tab:math_results} in the Appendix for comprehensive standardized results of recent RLVR models.}
\label{tab:gap_analysis}
\small
\begin{threeparttable}
\resizebox{\textwidth}{!}{%
\begin{tabular}{l|c|ccr}
\toprule
\textbf{Model (checkpoint)} &
\textbf{Benchmark} & \textbf{Reported (setup)} &
\textbf{Standardized Eval} & \textbf{$\Delta$} \\
\midrule
  & AIME-24   & 48.13 (\textit{pass@1}; \citealt{liu2025prorl}) & 45.62 & $+2.51$ \\
  & AIME-25   & 33.33 (\textit{pass@1}; \citealt{liu2025prorl}) & 33.85 & $-0.52$ \\
 Nemotron-Research- & AMC-23    & 79.29 (\textit{pass@1}; \citealt{liu2025prorl}) & 85.70 & $-6.41$ \\
 Reasoning-Qwen-1.5B v1 & Math      & 91.89 (\textit{pass@1}; \citealt{liu2025prorl}) & 92.01 & $-0.12$ \\
  & Minerva   & 47.98 (\textit{pass@1}; \citealt{liu2025prorl}) & 39.27 & $+8.71$ \\
  & Olympiad  & 60.22 (\textit{pass@1}; \citealt{liu2025prorl}) & 64.56 & $-4.34$ \\
\midrule
\multirow{2}{*}{AceReason-Nemotron-14B}
  & AIME-24   & 78.60 (\textit{avg@64}; \citealt{liu2025acereason}) & 77.29 & $+1.31$ \\
  & AIME-25   & 67.40 (\textit{avg@64}; \citealt{liu2025acereason}) & 66.04 & $+1.36$ \\
\midrule
DAPO-Qwen-32B
  & AIME-24   & 50.00 (\textit{avg@32}; $T{=}1.0$, top-$p{=}0.7$; \citealt{yu2025dapo}) & 51.56 & $-1.56$ \\
\midrule
Open-RS3-1.5B
  & AIME-24   & 46.70 (\textit{pass@1}; \citealt{dang2025reinforcement}) & 30.94 & $+15.76$ \\
STILL-3-1.5B
  & AIME-24   & 39.33 (\textit{pass@1}/accuracy; 5 sampled responses; \citealt{still3_tr}) & 31.46 & $+7.87$ \\
DeepScaleR-1.5B
  & AIME-24   & 43.10 (\textit{pass@1}; \citealt{deepscaler2025}) & 38.54 & $+4.56$ \\
\midrule
\multirow{2}{*}{Polaris-7B-Preview}
  & AIME-24   & 72.60 (\textit{avg@32}; \citealt{polaris_hf_card}) & 66.46 & $+6.14$ \\
  & AMC-23    & 89.00 (\textit{avg@8}; \citealt{polaris_hf_card}) & 93.59 & $-4.59$ \\
\bottomrule
\end{tabular}}
\end{threeparttable}
\end{table*}

\subsection{Reported vs.\ Reproduced: a Gap Analysis}
To make these issues concrete, we compared widely cited checkpoints to budget-parity controlled runs using the same verifier, matched decoding budgets, and aligned dataset versions. 

\textbf{Important:} some sources report \textit{pass@1} while others report \textit{avg@k}, sampled accuracy, or alternative aggregation under different decoding settings; in those cases, $\Delta$ conflates metric and budget differences with genuine policy differences. We therefore treat Table~\ref{tab:gap_analysis} as evidence of evaluation sensitivity. Furthermore, the gaps in Table~\ref{tab:gap_analysis} are exacerbated by three main factors:

\paragraph{Sampling budgets.} Multi-sample reporting and prompt/decoding choices can materially change reported scores. For example, AceReason-Nemotron-14B reports \textit{avg@64}, DAPO-Qwen-32B reports \textit{avg@32}, and STILL-3-1.5B reports 39.33 on AIME-24 under its own sampled \textit{pass@1} evaluation~\citep{liu2025acereason,yu2025dapo,still3_tr}. In our standardized runs, this manifests as $+7.87$ for STILL and $+6.14$ for Polaris-7B on AIME-24~\citep{polaris_hf_card}.

\paragraph{Template and version drift.} Changes in prompt templates and dataset slices (e.g., AIME curation, AMC answer formats) shift accuracy by several points, with exact hashes frequently missing from model cards~\citep{polaris_hf_card,openrs_hf_card}. This helps explain mixed signs for Nemotron-1.5B v1 (Minerva $+8.71$ vs.\ AMC-23 $-6.41$).

\paragraph{Metric and decision rules.} Sources sometimes mix ``accuracy,'' \textit{pass@k}, \textit{avg@k} (mean of \textit{pass@1} across draws), and \textit{maj@k} (majority vote), or substitute judges where verifiers exist. Once we normalize to a single verifier/spec, several deltas compress (e.g., DAPO-32B on AIME-24: $-1.56$)~\citep{sun2025evaluation,zhao2025onetoken}.

Finally, small-set variance matters: on AIME-24/25, seed and decode settings alone can yield $\pm$3--5 percentage points. Confidence intervals and saturation curves are needed to separate noise from effect~\citep{mu2025dissecting,hochlehnert2025sober}.

\paragraph{Why this matters}
Under unmatched budgets, unprobed judges, and untracked calibration, distributional \emph{sharpening} can masquerade as capability \emph{expansion}. Clean, parity-controlled evaluation often shrinks or flips celebrated gains. Hence, our minimum standard (budget parity and saturation curves, variance disclosure, judge robustness, calibration metrics, and a contamination audit in \secref{sec:contamination}) is necessary to distinguish real gains from artifacts.

\section{Perspective III: Data Contamination}
\label{sec:contamination}

Data provenance is a first-order confound: if pretraining or RL data overlap evaluation sets, measured ``reasoning'' may reflect memorization. Contamination-aware evaluations in coding~\citep{liu2024livecodebench} and multi-domain leaderboards~\citep{liang2023helm} already recommend strict provenance checks; our partial-prompt audit provides direct evidence in math. Partial-prompt reconstruction is a high-precision but not exhaustive signal of overlap: we treat it as one useful probe to be triangulated with lexical and fuzzy matching and with fresh test sets.

\begin{table}[!t]
\centering
\small
\caption{Partial-prompt contamination summary. The Math block (left) reports answer-match accuracy at an 80\% prefix on two legacy sets vs.\ the fresh \textsc{AIME-2025}. The SimpleQA block (right) is a non-math control, where \textsc{Qwen}'s advantage largely attenuates. Full results, including ACC@60/40 and ROUGE-L/EM for all models and datasets, are in Tables~\ref{tab:contam_qwen}--\ref{tab:simpleqa_contam} in the Appendix.}
\label{tab:contam_summary}
\resizebox{\linewidth}{!}{%
\begin{tabular}{l|cc|cc}
\toprule
\multirow{2}{*}{\textbf{Model}} &
\multicolumn{2}{c|}{\textbf{Math (ACC@80)}} &
\multicolumn{2}{c}{\textbf{SimpleQA}} \\
 & MATH-500 & AIME-2025 & R@80 & EM@80 \\
\midrule
Llama-3-1.8B    &  2.8 & 0.0  & 37.11 & 19.86 \\
Qwen2.5-Math-7B & 58.0 & 0.0  & 29.34 & 12.47 \\
Qwen2.5-32B     & 60.0 & 3.3  & 41.06 & 23.09 \\
Qwen3-14B-Base  & 58.2 & 0.0  & 37.24 & 19.40 \\
\bottomrule
\end{tabular}}
\end{table}

\paragraph{Probe design (partial-prompt completion).}
Following \citet{Wu2025Reasoning}, we reveal only the first $x\%$ of each problem ($x\in\{80,60,40\}$), greedily decode the suffix, and compute ROUGE-L@{$x$}, EM@{$x$}, and ACC@{$x$}. High EM/ACC under large prefixes indicates the model can reconstruct the hidden tail and final answer verbatim.

\paragraph{Case study: \textsc{Qwen} vs.\ \textsc{Llama}.}
As summarized in Table~\ref{tab:contam_summary}, \textsc{Qwen2.5-Math-7B} and \textsc{Qwen3-14B-Base} achieve high ACC@80 on legacy math sets (around 58--60\% on \textsc{MATH-500}) yet collapse on \textsc{AIME-2025} (0--3\%). \textsc{Llama-3-1.8B} remains near zero across the same sets, consistent with clean evaluation. This pattern (strong on older math sets, absent on the newest release) is consistent with substantial overlap between legacy math benchmarks and the training corpora of some \textsc{Qwen}-family models, and with comparatively cleaner evaluation for \textsc{Llama}. (See Appendix Tables~\ref{tab:contam_qwen}--\ref{tab:contam_qwen25_rougel} for ACC/ROUGE/EM at 80/60/40\% prefixes.) Similar contamination-aware stance and methodology are advocated in code by \citet{liu2024livecodebench} and in broader evaluations by \citet{liang2023helm}.

\paragraph{SimpleQA control.}
On a non-math control (\textsc{SimpleQA}) there is no systematic \textsc{Qwen} advantage: for example, \textsc{Qwen2.5-32B} attains 41.06/23.09 (R@80/EM@80) versus \textsc{Llama-3-1.8B} at 37.11/19.86 (Appendix Table~\ref{tab:simpleqa_contam}). The attenuation on an unseen domain supports the interpretation that elevated partial-prompt math scores on legacy sets reflect contamination rather than a general suffix-reconstruction ability. We use SimpleQA as a non-math control to check that strong tail-reconstruction behavior does not appear uniformly across domains. While suffix reconstruction may be easier for templated math than for open-domain QA, the absence of a consistent advantage on SimpleQA suggests that the large Qwen deltas on legacy math are not explained by a domain-general completion skill alone.

\paragraph{Controls and implications.}
We (i) treat contaminated sets as \emph{probes} of memorization, not reasoning; (ii) prioritize uncontaminated or freshly released test sets; and (iii) publish prompts/seeds/configurations to enable third-party screening. When we re-evaluate RLVR vs.\ base models under these controls, widely cited gaps shrink or flip. This directly supports our thesis: contamination can make modest distributional sharpening appear as frontier expansion.

\section{The Tax-Aware Minimum Standard for RLVR}
\label{sec:protocol}

Our results in \secref{sec:evaluation} and \secref{sec:contamination} show that three pieces of evaluation controls materially change how RLVR gains appear: parity-controlled budgeting, calibration and attempt tracking, and contamination audits. Alongside synthesized prior work on judge robustness and safety evaluations, we package these into a \emph{tax-aware minimum standard} for reporting RLVR results. The standard is intentionally narrow and measurement-focused. It does not prescribe a particular RL algorithm. Instead, it specifies the controls that must be in place before headline reasoning gains are treated as reliable.

\paragraph{(1) Budget parity and saturation curves.}
First, RLVR models and their baselines must be evaluated under matched sampling budgets. In our gap analysis, several celebrated improvements shrank or disappeared once we fixed the verifier, prompt family, decoding parameters, and number of samples per item (Table~\ref{tab:gap_analysis}). Reporting \textit{pass@k} for the RLVR model and \textit{pass@1} for the base model, or silently changing temperature, top-$p$, or stopping rules, makes it impossible to tell how much of the gain comes from extra search rather than a better policy.

Under the minimum standard, any claim that RLVR improves a metric on a given benchmark must include: (i) the exact sampling budget for both base and RLVR models, (ii) a saturation curve that plots accuracy as a function of $k$ under a shared decoding setup, and (iii) at least three seeds with mean and confidence intervals or standard deviations~\citep{mu2025dissecting,hochlehnert2025sober}. We recommend summarizing performance by area under the saturation curve in addition to a single \textit{pass@k} point, since this is less sensitive to one particular choice of $k$ and reveals whether RLVR shifts the whole budget-performance frontier or only improves at very large $k$.

\paragraph{(2) Calibration, abstention, and judge robustness.}
Second, evaluations must track calibration and abstention, not only accuracy. \secref{sec:tax} showed that RLVR often reduces refusals and increases stated confidence, which yields more attempted items but also more confident errors. Our factual QA control (Table~\ref{tab:factuality-compact}) separates \emph{shared} accuracy from \emph{newly attempted} tail items and measures expected calibration error (ECE) on shared items. This reveals a distinct failure mode: headline scores rise because the model stops saying ``I do not know'' and starts answering everything, while accuracy on overlapping items barely moves.

Under the minimum standard, each reported accuracy figure must be accompanied by: (i) refusal or ``not attempted'' rates, (ii) shared accuracy on the intersection of items both models attempted, and (iii) a calibration metric such as ECE computed from the model's confidence scores~\citep{hochlehnert2025sober,leng2024taming}. For settings that rely on LLM-as-a-judge rather than programmatic verifiers, we also require at least one \emph{judge stress test}: the same outputs scored under several prompt templates or instruction orderings, with the spread in scores reported~\citep{zhao2025onetoken,sun2025evaluation,yang2026toward}. This does not remove judge fragility, but it makes visible how sensitive a claimed RLVR gain is to small changes in the judge pipeline. Together, these measurements ensure that gains are not driven primarily by attempt inflation, miscalibration, or a brittle judge configuration.

\paragraph{(3) Contamination audits and data hygiene.}
Third, RLVR claims must be supported by explicit data provenance checks. \secref{sec:contamination} used partial-prompt completion to show that some math benchmarks are heavily memorized by \textsc{Qwen} families: the model can reconstruct the hidden tail and final answer when given 80 percent of the problem on legacy sets, but not on fresh ones such as \textsc{AIME-2025}. Without such checks, apparent reasoning gains can simply reflect better recall of training data~\citep{liu2024livecodebench,liang2023helm,Wu2025Reasoning}.

Under the minimum standard, any benchmark used to support an RLVR claim must be accompanied by: (i) a contamination screen that combines fuzzy or lexical matching against pretraining and fine-tuning corpora (where available) and partial-prompt probes at several prefix lengths, and (ii) at least one clean held-out set that shows no evidence of tail reconstruction even at large prefixes. For benchmarks that do show contamination, we treat them as \emph{probes} of memorization rather than as primary evidence of reasoning and clearly label them as such. We also require that dataset versions, prompt templates, and filtering rules be released or described precisely enough for others to re-run the audit.

\paragraph{Summary.}
The tax-aware minimum standard asks for three things before treating RLVR gains as robust: matched budgets with saturation curves and variance disclosure, calibration and abstention metrics with at least one judge stress test when judges are used, and a contamination audit with at least one clean held-out benchmark. Taken together, these controls provide a simple, concrete bar that future RLVR work can meet without specifying a particular training recipe, and that can be reused for SFT, RLHF, and test-time compute evaluations.

\section{Conclusion}
Our central position is that RLVR is effective and deployable for verifiable domains, but headline ``reasoning'' gains should be treated as provisional unless a small core of tax-aware controls (budget parity, calibration/abstention tracking, robust evaluation with at least one judge stress test, and a simple contamination audit) is enforced. RLVR delivers real gains on verifiable tasks, but the field often over-indexes on headline accuracy while under-weighting \emph{taxes} and \emph{measurement}. Under budget parity controls and calibration tracking, several celebrated ``reasoning gains'' shrink, suggesting that part of the progress reflects distributional \emph{sharpening} rather than durable expansion.



\section*{Limitations}
Our goal is measurement clarity rather than a complete theory of when RLVR yields capability expansion, so our evidence has clear boundaries.
First, our empirical analyses focus on verifiable domains (math, code, structured QA) using open checkpoints and public benchmarks; findings may not transfer directly to agentic, multimodal, or tool-use settings where programmatic verification is harder.
Second, our attempt inflation and calibration analyses depend on operationalizing "attempts" (e.g., extractable answers versus explicit refusals); different extraction rules shift absolute rates, so we emphasize shared-item comparisons to reduce sensitivity.
Third, our partial-prompt contamination probe is a high-precision indicator of tail reconstruction on some datasets but is not exhaustive: it can miss semantic overlap and does not quantify training-set membership. We treat it as one triangulation signal alongside provenance checks and fresh held-out sets. We omit approximate nearest-neighbor or fuzzy-hash matching against training corpora and shard-level provenance analysis, which would strengthen contamination claims but require additional infrastructure and training-data access.
Finally, we exclude RLVR reward-component ablations (e.g., explicit refusal, grounding, or calibration terms) because they require additional training runs; we view them as promising next steps.

\section*{Ethical Considerations}
This paper is primarily about evaluation and reporting practices, but it touches on safety and privacy.
Longer reasoning traces can increase the chance of exposing sensitive strings or unsafe intermediate content.
To mitigate risk, we restrict experiments to public benchmarks and report aggregate statistics rather than
releasing jailbreak payloads or automated attack scripts.
We also highlight calibration, abstention, and provenance as first-class reporting targets because inflated headline accuracy can create misleading impressions of real-world reliability.
Finally, our recommendations can increase evaluation cost (e.g., multi-seed, budget-matched saturation curves); we view this as a necessary trade-off for measurement integrity when results are used to justify deployment.



\bibliography{cite}

\clearpage
\appendix

\section{Appendix}
\setlength{\tabcolsep}{5.5pt}
\renewcommand{\arraystretch}{1.08}

\paragraph{Conventions and metrics.}
\textit{pass@k} = probability at least one of $k$ samples is correct; \textit{avg@k} = mean \textit{pass@1} across $k$ draws; \textit{maj@k} = majority vote over $k$; \textit{ECE} = expected calibration error. Arrows ($\uparrow$/$\downarrow$) indicate whether higher or lower is better.

\subsection{Supplementary Figure for the Introduction}
\label{app:intro_figs}

\begin{figure*}[ht]
  \centering
  \includegraphics[width=\linewidth]{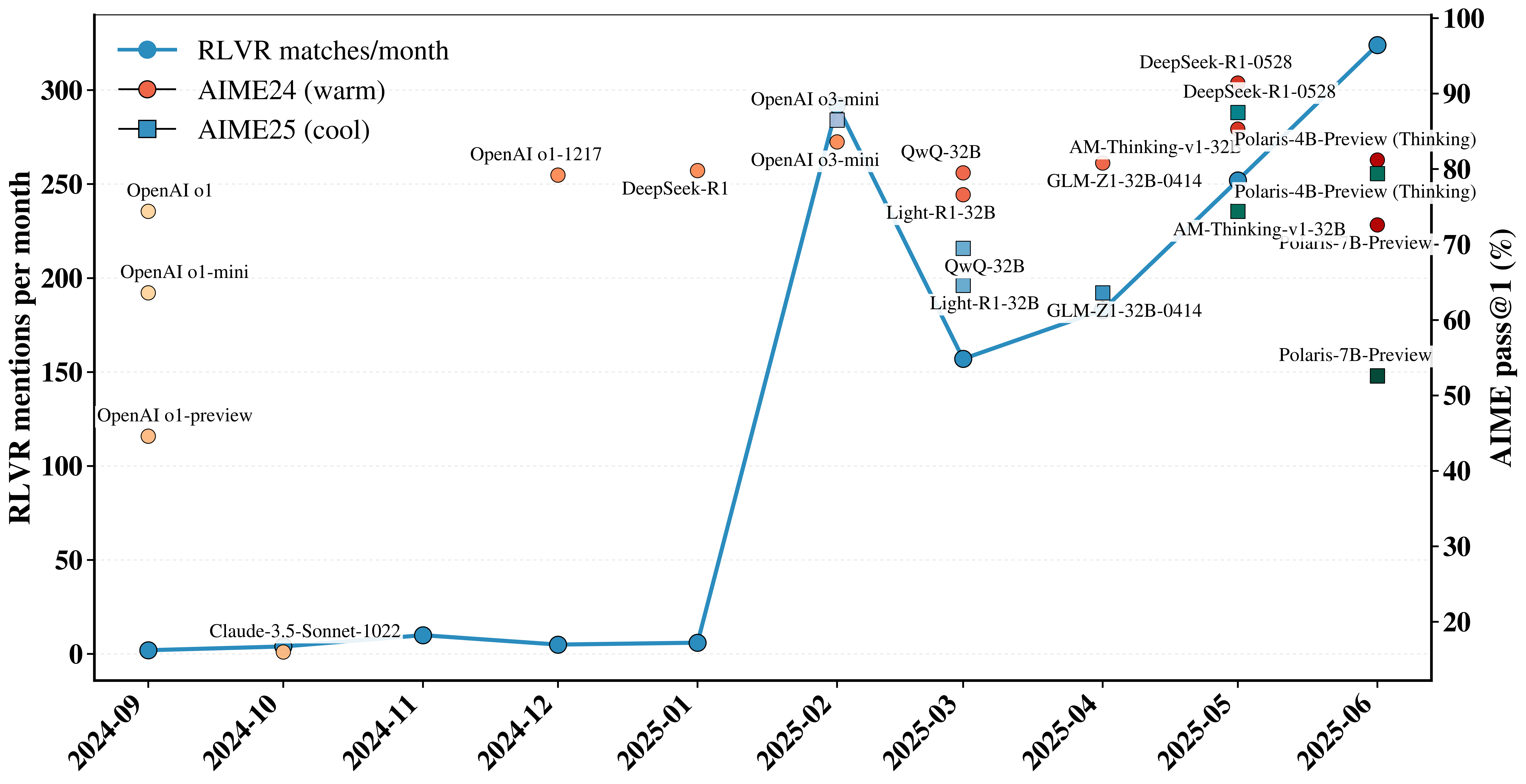}%
  \caption{Monthly RLVR activity vs.\ AIME performance (time span: May~2024--June~2025).
  \textbf{Left axis}: count of pages per month whose \emph{title or abstract} contains
  ``RLVR'' or ``reinforcement learning with verifiable rewards'' (Google Scholar and arXiv via SerpAPI).
  \textbf{Right axis}: \textit{pass@1} (\%) on AIME-24 (circles) and AIME-25 (squares) for selected models; labels show model names.}
  \label{fig:rlvr_trend}
\end{figure*}

\noindent This appendix subsection contains Fig.~\ref{fig:rlvr_trend}, which shows the RLVR activity trend alongside AIME performance.

\subsection{Potential Criticisms and Responses}
\label{app:criticisms}

\paragraph{On capability expansion.}
RLVR can expand capability under specific regimes (e.g., prolonged RL with resets and KL control; unlikeliness-based shaping; domain-targeted RLVR with sparse pretraining)~\citep{liu2025prorl,he2025rewarding,cheng2025revisiting}. Our claim is not negation but \emph{measurement}: without tax-aware training and parity-controlled evaluation, expansion is often overstated. The protocol in \secref{sec:protocol} is compatible with these positive regimes while curbing overclaiming.

\paragraph{On ``it's just data.''}
Data quality and curricula matter~\citep{chen2025bridging,wen2025light}, but reward/metric design independently shapes failure modes (determinacy/overconfidence under judge rewards; terse/off-topic outputs under naive factuality; calibration drift)~\citep{chen2025learningreasonfactuality,leng2024taming,hochlehnert2025sober}. We treat \emph{both} data and objectives as first-class knobs (see \secref{sec:protocol}).

\paragraph{On budgeted \textit{pass@k}.}
We use \textit{pass@k} for practical relevance, but ask for \emph{matched budgets} and \emph{saturation curves}. Under these controls, headline gaps often shrink or flip~\citep{hochlehnert2025sober}. Reporting AUC alongside \textit{pass@k} keeps comparisons fair (\secref{sec:evaluation}).

\paragraph{On safety/privacy risk.}
Jailbreak success scales with attempts/context; many-shot tuning can reduce measured rates in controlled settings~\citep{zhou2025hidden,jiang2025safechain,ackerman2025manyshot}. Longer CoT increases leakage surfaces~\citep{green2025leakythoughtslargereasoning,zhou2025trust}. Our protocol caps Best-of-$N$/CoT length at eval and co-optimizes abstention/privacy with accuracy (\secref{sec:protocol}).

\paragraph{On reliance on LLM judges.}
When verifiers are unavailable or partial, we use LLM judges with robustness probes (prompt/format perturbations, adversarial tests) and publish templates/configs; judges are manipulable, so robustness deltas are reported~\citep{zhao2025onetoken,sun2025evaluation}. Prefer verifiers when possible (\secref{sec:evaluation}).

\paragraph{Scope.}
Our synthesis centers on verifiable math/code/QA with open-weight models. Agentic/multimodal settings introduce additional privacy/safety channels and may need tailored verifiers and audits; we expect core principles (budget parity, calibration, abstention, contamination screens) to transfer with domain-specific adaptations.

\subsection{Standardized Evaluations}
\label{app:standard}

We report full results under matched budgets, fixed verifiers, and a shared prompt family.
These tables extend \secref{sec:evaluation} and underlie the saturation-curve argument: once $k$, templates, and dataset versions are controlled, several headline gaps narrow.
\textbf{See Table~\ref{tab:math_results} below for the full standardized benchmark results.}
All results in this section use the same verifier and prompt template family; we specify $k$, number of seeds, and dataset versions below.

\paragraph{Key observations.}
Thinking-mode (test-time compute) delivers large gains across sizes: for 4B models, Qwen3-4B$\to$Qwen3-4B (Thinking) yields +50.5/+44.7 on AIME-24/25 and $\approx$+28 on the averaged score; Polaris-4B-Preview$\to$Thinking shows +52.6/+54.7 and $\approx$+29 on Avg; for 8B, Qwen3-8B$\to$Thinking adds +51.9/+46.3 and $\approx$+28 on Avg. Small RLVR-tuned models can exceed larger non-thinking baselines: the 1.5B Nemotron-Research-Reasoning v2 averages \textbf{61.70}, outscoring non-thinking 4--8B baselines (Qwen3-4B \textbf{46.98}, Qwen3-8B \textbf{48.96}) and leading them by $\sim$24--30 points on AIME-24. Within size families, leaders are consistent across benchmarks (7B AceReason-1.1 Avg \textbf{74.44}; 14B AceReason Avg \textbf{77.33}; 32B DeepSeek-R1-Distill Avg \textbf{72.53}), while the frontier DeepSeek-R1-0528 tops overall (AIME-24 \textbf{90.00}, AIME-25 \textbf{78.33}, Avg \textbf{81.72}). Several columns (AMC/MATH) cluster in the mid-90s, indicating saturation under our $k{=}32$/verifier setup. Together, these patterns support (i) test-time compute as an effective lever for reasoning; (ii) RLVR's disproportionate lift at smaller scales; and (iii) the need for matched budgets, response-length controls, and saturation curves to interpret small nominal deltas fairly.

\begin{table*}[!htbp]
\centering
\caption{Standardized evaluation across math benchmarks (higher is better). \emph{Styling:} within each size family, the best score for a given benchmark is \textbf{bold} and the second-best is \underline{underlined}. \textbf{Context:} Parity-controlled scores used in \secref{sec:evaluation}. \textbf{Setup:} \textit{pass@1} estimated by averaging over $k{=}32$ independent single-sample decodes (\textit{avg@32}), same verifier and prompt family, 3 seeds. \textbf{Compute:} evaluations ran on a $128{\times}$H100 (96\,GB) cluster and consumed $\approx$3{,}500 GPU-hours.}
\label{tab:math_results}
\resizebox{\textwidth}{!}{%
\begin{tabular}{l|cccccc|c}
\toprule
\textbf{Model} & \textbf{AIME-24}$\uparrow$ & \textbf{AIME-25}$\uparrow$ & \textbf{AMC-23}$\uparrow$ & \textbf{MATH}$\uparrow$ & \textbf{Minerva}$\uparrow$ & \textbf{Olympiad}$\uparrow$ & \textbf{Avg}$\uparrow$ \\
\midrule
Qwen2.5-Math-1.5B & 8.33 & 6.35 & 44.06 & 66.67 & 18.42 & 30.74 & 29.10 \\
Qwen2.5-Math-1.5B-Instruct & 10.10 & 8.85 & 55.08 & 74.83 & 29.32 & 40.00 & 36.37 \\
DeepSeek-R1-Distill-Qwen-1.5B & 31.15 & 24.06 & 72.81 & 85.01 & 32.18 & 51.55 & 49.46 \\
STILL-3-1.5B & 31.46 & 25.00 & 75.08 & 86.24 & 32.77 & 53.84 & 50.73 \\
DeepScaleR-1.5B & 38.54 & 30.52 & 80.86 & 88.79 & 36.19 & 58.95 & 55.64 \\
Qwen2.5-Math-1.5B-Oat-Zero & 20.00 & 10.00 & 52.50 & 74.20 & 26.84 & 37.78 & 36.89 \\
Open-RS1 & 30.94 & 22.60 & 73.05 & 84.90 & 29.92 & 52.82 & 49.04 \\
Open-RS2 & 28.96 & 24.37 & 73.52 & 85.06 & 29.74 & 52.63 & 49.05 \\
Open-RS3 & 30.94 & 24.79 & 72.50 & 84.47 & 29.11 & 52.25 & 49.01 \\
Nemotron-Research-Reasoning-Qwen-1.5B v1 & \secondbest{45.62} & \best{33.85} & \secondbest{85.70} & \secondbest{92.01} & \secondbest{39.27} & \secondbest{64.56} & \secondbest{60.17} \\
Nemotron-Research-Reasoning-Qwen-1.5B v2 & \best{51.77} & \secondbest{32.92} & \best{88.83} & \best{92.24} & \best{39.75} & \best{64.69} & \best{61.70} \\
\midrule
Qwen2.5-Math-7B & 15.62 & 6.56 & 52.81 & 67.72 & 15.64 & 32.44 & 31.80 \\
Qwen2.5-Math-7B-Instruct & 12.19 & 9.17 & 58.36 & 83.21 & 35.56 & 41.60 & 40.01 \\
DeepSeek-R1-Distill-Qwen-7B & 53.23 & 38.96 & 89.30 & 93.95 & 43.07 & 66.67 & 64.20 \\
Qwen2.5-Math-7B-Oat-Zero & 26.67 & 6.67 & 67.50 & 79.20 & 32.72 & 41.78 & 42.42 \\
Skywork-OR1-7B & \secondbest{66.88} & 51.15 & 92.73 & \secondbest{96.04} & 44.03 & 73.61 & 70.74 \\
LEAD-7B & 51.67 & 37.19 & 89.06 & 93.73 & 43.11 & 66.34 & 63.52 \\
AceReason-Nemotron-7B & 65.83 & 47.19 & \best{95.08} & 95.81 & \best{45.35} & \secondbest{73.92} & 70.53 \\
AceReason-Nemotron-1.1-7B & \best{71.56} & \best{64.58} & 93.36 & \best{96.73} & 44.05 & \best{76.37} & \best{74.44} \\
Polaris-7B-Preview & 66.46 & \secondbest{51.56} & \secondbest{93.59} & 95.68 & \secondbest{44.47} & 73.65 & \secondbest{70.90} \\
\midrule
Qwen2.5-14B & 11.04 & 7.92 & 47.19 & 73.19 & 22.51 & 37.01 & 33.14 \\
Qwen2.5-14B-Instruct & 13.65 & 12.40 & 58.13 & 80.28 & 38.63 & 43.23 & 41.05 \\
DeepSeek-R1-Distill-Qwen-14B & \secondbest{67.81} & 48.33 & \secondbest{95.39} & \secondbest{95.28} & 46.43 & 72.06 & \secondbest{70.88} \\
LEAD-14B & 64.06 & \secondbest{52.29} & 92.81 & 95.23 & \secondbest{47.52} & \secondbest{72.25} & 70.69 \\
AceReason-Nemotron-14B & \best{77.29} & \best{66.04} & \best{98.67} & \best{96.90} & \best{47.73} & \best{77.34} & \best{77.33} \\
\midrule
Qwen2.5-32B & 15.62 & 9.17 & 59.30 & 76.51 & 26.42 & 41.45 & 38.08 \\
Qwen2.5-32B-Instruct & 17.19 & 14.17 & 67.66 & 83.17 & 40.96 & 47.85 & 45.17 \\
DeepSeek-R1-Distill-Qwen-32B & \best{69.06} & \best{55.52} & \best{95.62} & \best{95.74} & \best{46.50} & \best{72.76} & \best{72.53} \\
DAPO-Qwen-32B & 51.56 & 36.98 & \secondbest{92.73} & 80.74 & 33.07 & 48.97 & 57.34 \\
Enigmata-Qwen2.5-32B & \secondbest{61.67} & \secondbest{46.88} & 91.25 & \secondbest{93.69} & \secondbest{46.32} & \secondbest{69.14} & \secondbest{68.16} \\
\midrule
DeepSeek-V3-0324 & \secondbest{55.83} & \secondbest{43.33} & \secondbest{92.50} & \secondbest{95.12} & \best{48.58} & \secondbest{66.91} & \secondbest{67.05} \\
DeepSeek-R1-0528 & \best{90.00} & \best{78.33} & \best{99.38} & \best{97.80} & \secondbest{48.35} & \best{76.44} & \best{81.72} \\
\midrule
Qwen3-4B & 21.88 & 17.92 & 66.95 & 84.27 & 38.50 & 52.38 & 46.98 \\
Qwen3-4B (Thinking) & \secondbest{72.40} & \secondbest{62.60} & \secondbest{95.78} & \secondbest{96.31} & \secondbest{46.44} & \secondbest{76.40} & \secondbest{74.99} \\
Polaris-4B-Preview & 27.29 & 23.02 & 71.56 & 85.89 & 39.20 & 59.13 & 51.02 \\
Polaris-4B-Preview (Thinking) & \best{79.90} & \best{77.71} & \best{99.45} & \best{97.38} & \best{47.21} & \best{80.38} & \best{80.34} \\
\midrule
Qwen3-8B & 25.42 & 20.21 & 69.61 & 84.54 & 39.81 & 54.15 & 48.96 \\
Qwen3-8B (Thinking) & \best{77.29} & \secondbest{66.46} & \secondbest{95.00} & \best{96.86} & \best{49.06} & \best{77.56} & \best{77.04} \\
DeepSeek-R1-0528-Qwen3-8B & \secondbest{75.73} & \best{67.29} & \best{97.19} & \secondbest{96.32} & \secondbest{45.85} & \secondbest{74.19} & \secondbest{76.10} \\
\bottomrule
\end{tabular}}
\end{table*}

\subsection{Sequence Length and Compute Footprint}
\label{app:token}

We summarize generated token lengths by benchmark as a proxy for compute footprint and exposure surface.
Longer chains increase latency and may worsen calibration and safety/privacy risks (\secref{sec:tax}),
so we report typical generation lengths under the same decoding budgets used in \appref{app:standard}.
\textbf{See Table~\ref{tab:token_length_results} below for the token-length breakdown by model and benchmark.}

\begin{table*}[!htbp]
\centering
\caption{Median generated output length (tokens per problem; answer + reasoning) under the standardized decoding setup.
\textbf{Context:} Compute/privacy footprint complementing accuracy.
\textbf{Setup:} Same prompts/verifier as \appref{app:standard}; decoding budget $k{=}32$; lengths aggregated across all samples per problem (correct and incorrect).
\textbf{Takeaway:} Reasoning-tuned models often produce much longer traces (e.g., 2--8$\times$ vs.\ instruct baselines), which impacts evaluation cost and safety/privacy exposure.}
\label{tab:token_length_results}
\resizebox{\textwidth}{!}{%
\begin{tabular}{l|cccccc|c}
\toprule
\textbf{Model} & \textbf{AIME-24} & \textbf{AIME-25} & \textbf{AMC-23} & \textbf{Math} & \textbf{Minerva} & \textbf{Olympiad} & \textbf{Avg} \\
\midrule
Qwen2.5-Math-1.5B & 1114 & 1033 & 828 & 648 & 959 & 904 & 914 \\
Qwen2.5-Math-1.5B-Instruct & 990 & 891 & 801 & 566 & 640 & 810 & 783 \\
DeepSeek-R1-Distill-Qwen-1.5B & 16363 & 16252 & 9979 & 5700 & 8194 & 11873 & 11394 \\
STILL-3-1.5B & 13350 & 13000 & 7716 & 4314 & 5921 & 9345 & 8941 \\
DeepScaleR-1.5B & 9780 & 8978 & 5003 & 3139 & 5270 & 5807 & 6330 \\
Qwen2.5-Math-1.5B-Oat-Zero & 1166 & 1155 & 848 & 626 & 689 & 892 & 896 \\
Open-RS1 & 13578 & 13698 & 7495 & 4170 & 5935 & 8995 & 8978 \\
Open-RS2 & 14215 & 13618 & 7612 & 4174 & 5805 & 9059 & 9081 \\
Open-RS3 & 14394 & 13606 & 7884 & 4233 & 5688 & 9047 & 9142 \\
Nemotron-Research-Reasoning-Qwen-1.5B & 7786 & 7713 & 6294 & 5070 & 6569 & 6678 & 6685 \\
\midrule
Qwen2.5-Math-7B & 1201 & 1152 & 946 & 720 & 1136 & 950 & 1018 \\
Qwen2.5-Math-7B-Instruct & 1465 & 1418 & 983 & 670 & 748 & 1051 & 1056 \\
DeepSeek-R1-Distill-Qwen-7B & 13613 & 14543 & 6402 & 4125 & 5595 & 8988 & 8878 \\
Qwen2.5-Math-7B-Oat-Zero & 1032 & 1178 & 884 & 673 & 696 & 870 & 889 \\
Skywork-OR1-7B & 15366 & 17845 & 8361 & 5541 & 8566 & 11818 & 11250 \\
LEAD-7B & 10838 & 11573 & 4863 & 3111 & 3692 & 7054 & 6855 \\
AceReason-Nemotron-1.1-7B & 14331 & 16502 & 6672 & 3835 & 6676 & 9060 & 9513 \\
Polaris-7B-Preview & 12564 & 14389 & 6538 & 4313 & 6125 & 8681 & 8768 \\
\midrule
Qwen2.5-14B & 1076 & 1088 & 847 & 591 & 1209 & 815 & 938 \\
Qwen2.5-14B-Instruct & 1079 & 994 & 850 & 607 & 632 & 844 & 834 \\
DeepSeek-R1-Distill-Qwen-14B & 11295 & 13389 & 5735 & 3781 & 4919 & 8042 & 7860 \\
LEAD-14B & 8364 & 9019 & 4469 & 3031 & 4675 & 5713 & 5879 \\
AceReason-Nemotron-14B & 13871 & 16334 & 7239 & 4609 & 7677 & 10030 & 9960 \\
\midrule
Qwen2.5-32B & 1165 & 1055 & 782 & 556 & 1186 & 800 & 924 \\
DeepSeek-R1-Distill-Qwen-32B & 10979 & 13012 & 5826 & 3652 & 4663 & 7924 & 7676 \\
DAPO-Qwen-32B & 6627 & 6074 & 3086 & 2812 & 3887 & 5122 & 4601 \\
Enigmata-Qwen2.5-32B & 13204 & 15417 & 10195 & 6358 & 13770 & 12098 & 11840 \\
\midrule
DeepSeek-V3-0324 & 3601 & 3474 & 2134 & 1281 & 705 & 2386 & 2264 \\
DeepSeek-R1-0528 & 16276 & 18416 & 10285 & 6041 & 7085 & 13750 & 11976 \\
\midrule
MiMo-7B-Base & 21716 & 18842 & 15734 & 6390 & 13809 & 13299 & 14965 \\
MiMo-7B-SFT & 13012 & 14250 & 7030 & 4177 & 6152 & 8421 & 8841 \\
MiMo-7B-RL-Zero & 18105 & 18575 & 9950 & 6564 & 6546 & 12152 & 11982 \\
\midrule
Qwen3-4B & 8909 & 6079 & 2369 & 1289 & 812 & 3217 & 3779 \\
Qwen3-4B (Thinking) & 14750 & 17813 & 8290 & 5076 & 6682 & 10606 & 10536 \\
Polaris-4B-Preview & 5582 & 6148 & 2637 & 1293 & 804 & 4632 & 3516 \\
Polaris-4B-Preview (Thinking) & 28369 & 33109 & 15125 & 9333 & 13307 & 21832 & 20179 \\
\midrule
Qwen3-8B & 8375 & 6151 & 2863 & 1479 & 662 & 3624 & 3859 \\
Qwen3-8B (Thinking) & 14764 & 18296 & 8756 & 5391 & 7105 & 11384 & 10949 \\
DeepSeek-R1-0528-Qwen3-8B & 20970 & 22742 & 12895 & 7640 & 9329 & 16108 & 14947 \\
\bottomrule
\end{tabular}}
\end{table*}

\subsection{Data Contamination Probes: Math}
\label{app:contam-math}

We use partial-prompt completion to detect tail reconstruction: reveal the first $x\%$ of the problem ($x\in\{80,60,40\}$), greedily decode the remainder, and score ACC/EM/ROUGE-L at that prefix.
High scores at large prefixes indicate memorization of legacy sets rather than reasoning. Tables here expand \secref{sec:contamination}.
\textbf{Accuracy results are reported in Tables~\ref{tab:contam_qwen} and \ref{tab:contam_qwen25}; text-overlap and exact-match proxies are reported in Tables~\ref{tab:contam_qwen3_rougel} and \ref{tab:contam_qwen25_rougel}.}

\begin{table*}[t]
\centering
\small
\caption{Accuracy of \textsc{Qwen3} checkpoints on five mathematics benchmarks when each model receives only the first $x\%$ of the question ($x=80,60,40$) and must greedily complete the remainder. Columns ``ACC ($x\%$)'' report the average accuracy at that prefix length. \textbf{Takeaway:} \textsc{Qwen3} variants achieve high ACC@80 on legacy \textsc{MATH-500}/\textsc{AMC-23} but collapse on \textsc{AIME-2025}, consistent with contamination in older sets.}
\label{tab:contam_qwen}

\begin{adjustbox}{max width=\textwidth,center}
\begin{tabular}{l|l|ccc}
\toprule
\textbf{Model} & \textbf{Dataset} & \textbf{ACC (80\%)} & \textbf{ACC (60\%)} & \textbf{ACC (40\%)} \\
\midrule
                & MATH-500        & 26.40 & 14.40 &  5.40 \\
                & AMC-23          & 20.00 &  5.00 &  0.00 \\
Qwen3-0.6B-Base & AIME 2024       &  3.33 &  0.00 &  0.00 \\
                & AIME 2025       &  0.00 &  0.00 &  0.00 \\
                & Minerva-Math    &  1.10 &  0.00 &  0.00 \\
\midrule
                & MATH-500        & 33.80 & 18.40 &  9.00 \\
                & AMC-23          & 20.00 & 15.00 &  0.00 \\
Qwen3-1.7B-Base & AIME 2024       &  0.00 &  0.00 &  0.00 \\
                & AIME 2025       &  3.33 &  3.33 &  0.00 \\
                & Minerva-Math    &  2.94 &  2.21 &  0.74 \\
\midrule
                & MATH-500        & 43.20 & 28.20 & 15.00 \\
                & AMC-23          & 37.50 & 37.50 & 17.50 \\
Qwen3-4B-Base   & AIME 2024       & 10.00 &  6.67 & 10.00 \\
                & AIME 2025       &  3.33 &  0.00 &  0.00 \\
                & Minerva-Math    &  4.78 &  2.31 &  0.74 \\
\midrule
                & MATH-500        & 52.00 & 35.60 & 24.80 \\
                & AMC-23          & 35.00 & 25.00 & 27.50 \\
Qwen3-8B-Base   & AIME 2024       & 13.33 &  6.67 & 10.00 \\
                & AIME 2025       &  3.33 &  3.33 &  0.00 \\
                & Minerva-Math    &  5.88 &  3.31 &  1.84 \\
\midrule
                & MATH-500        & 58.20 & 44.80 & 30.20 \\
                & AMC-23          & 47.50 & 37.50 & 32.50 \\
Qwen3-14B-Base  & AIME 2024       & 10.00 & 16.67 & 16.67 \\
                & AIME 2025       &  0.00 &  0.00 &  0.00 \\
                & Minerva-Math    &  5.88 &  2.21 &  2.94 \\
\midrule
                & MATH-500        & 62.40 & 51.40 & 37.80 \\
                & AMC-23          & 40.00 & 25.00 & 25.00 \\
Qwen3-30B-A3B   & AIME 2024       & 13.33 & 23.33 & 13.33 \\
                & AIME 2025       &  3.33 &  0.00 &  0.00 \\
                & Minerva-Math    &  4.78 &  3.21 &  1.84 \\
\bottomrule
\end{tabular}
\end{adjustbox}

\end{table*}

\begin{table*}[t]
\centering
\small
\caption{Accuracy of \textsc{Qwen2.5} and \textsc{Llama-3} checkpoints on five mathematics benchmarks when each model receives only the first $x\%$ of the question ($x=80,60,40$) and must greedily complete the remainder. \textbf{Takeaway:} \textsc{Qwen2.5} shows strong tail reconstruction on legacy math; \textsc{Llama-3-1.8B} remains near zero across sets, reinforcing the contamination interpretation.}
\label{tab:contam_qwen25}

\begin{adjustbox}{max width=\textwidth,center}
\begin{tabular}{l|l|ccc}
\toprule
\textbf{Model} & \textbf{Dataset} & \textbf{ACC (80\%)} & \textbf{ACC (60\%)} & \textbf{ACC (40\%)} \\
\midrule
                 & MATH-500        & 58.00 & 44.40 & 29.20 \\
                 & AMC-23          & 52.50 & 42.50 & 32.50 \\
Qwen2.5-Math-7B  & AIME 2024       & 16.67 & 20.00 & 16.67 \\
                 & AIME 2025       &  0.00 &  0.00 &  0.00 \\
                 & Minerva-Math    &  7.72 &  4.41 &  2.94 \\
\midrule
                 & MATH-500        & 44.80 & 27.80 & 15.00 \\
                 & AMC-23          & 27.50 & 27.50 & 22.50 \\
Qwen2.5-7B       & AIME 2024       &  6.67 &  0.00 &  3.33 \\
                 & AIME 2025       &  3.33 &  3.33 &  0.00 \\
                 & Minerva-Math    &  6.62 &  4.78 &  1.47 \\
\midrule
                    & MATH-500     & 44.80 & 28.60 & 14.00 \\
                    & AMC-23       & 40.00 & 15.00 & 12.50 \\
Qwen2.5-7B-Instruct & AIME 2024    &  3.33 &  0.00 &  0.00 \\
                    & AIME 2025    &  3.33 &  0.00 &  0.00 \\
                    & Minerva-Math & 10.29 &  5.51 &  3.31 \\
\midrule
                 & MATH-500        & 50.60 & 35.80 & 21.00 \\
                 & AMC-23          & 40.00 & 27.50 & 27.50 \\
Qwen2.5-14B      & AIME 2024       & 10.00 &  3.33 &  6.67 \\
                 & AIME 2025       &  3.33 &  0.00 &  0.00 \\
                 & Minerva-Math    &  8.46 &  6.25 &  2.21 \\
\midrule
                 & MATH-500        & 60.00 & 47.80 & 32.00 \\
                 & AMC-23          & 52.50 & 45.00 & 42.50 \\
Qwen2.5-32B      & AIME 2024       & 16.67 & 13.33 & 10.00 \\
                 & AIME 2025       &  3.33 &  0.00 &  6.67 \\
                 & Minerva-Math    &  9.93 &  4.41 &  2.94 \\
\midrule
                 & MATH-500        &  2.80 &  2.40 &  2.00 \\
                 & AMC-23          &  0.00 &  0.00 &  0.00 \\
Llama-3-1.8B     & AIME 2024       &  0.00 &  0.00 &  0.00 \\
                 & AIME 2025       &  0.00 &  0.00 &  0.00 \\
                 & Minerva-Math    &  1.84 &  0.37 &  0.00 \\
\bottomrule
\end{tabular}
\end{adjustbox}

\end{table*}

\begin{table*}[t]
\centering
\scriptsize
\caption{RougeL and exact-match (EM) scores for \textsc{Qwen3.0} checkpoints on five mathematics benchmarks when each model receives only the first $x\%$ of the question ($x=80,60,40$) and must greedily complete the remainder. Columns ``R@x'' and ``EM@x'' report the average ROUGE-L and EM for that prefix length. \textbf{Takeaway:} Elevated ROUGE-L/EM at large prefixes on legacy sets (but not on \textsc{AIME-2025}) indicates suffix reconstruction rather than emergent reasoning.}
\label{tab:contam_qwen3_rougel}
\resizebox{\textwidth}{!}{%
\begin{tabular}{l|l|rrrrrr}
\toprule
\textbf{Model} & \textbf{Dataset} &
\multicolumn{1}{c}{\textbf{R@80}} & \multicolumn{1}{c}{\textbf{EM@80}} &
\multicolumn{1}{c}{\textbf{R@60}} & \multicolumn{1}{c}{\textbf{EM@60}} &
\multicolumn{1}{c}{\textbf{R@40}} & \multicolumn{1}{c}{\textbf{EM@40}} \\
\midrule
                & MATH-500        & 53.31 & 20.80 & 45.90 &  6.40 & 39.33 &  1.00 \\
                & AMC-23          & 52.26 &  7.50 & 40.39 &  0.00 & 33.92 &  0.00 \\
Qwen3-0.6B-Base & AIME 2024       & 54.85 & 20.00 & 27.79 &  0.00 & 24.76 &  0.00 \\
                & AIME 2025       & 54.69 & 10.00 & 33.47 &  0.00 & 29.16 &  0.00 \\
                & Minerva-Math    & 31.41 &  1.47 & 29.17 &  0.00 & 24.86 &  0.00 \\
\midrule
                & MATH-500        & 55.86 & 24.60 & 49.05 & 10.20 & 41.91 &  2.40 \\
                & AMC-23          & 64.35 & 30.00 & 52.57 & 20.00 & 44.95 & 12.50 \\
Qwen3-1.7B-Base & AIME 2024       & 56.55 & 26.67 & 36.48 &  6.67 & 36.56 &  6.67 \\
                & AIME 2025       & 53.90 & 16.67 & 39.40 &  6.00 & 31.09 &  0.00 \\
                & Minerva-Math    & 33.99 &  3.31 & 30.32 &  0.00 & 27.28 &  0.00 \\
\midrule
                & MATH-500        & 65.04 & 37.40 & 55.98 & 18.80 & 46.44 &  7.00 \\
                & AMC-23          & 73.24 & 52.50 & 68.56 & 45.00 & 63.89 & 35.00 \\
Qwen3-4B-Base   & AIME 2024       & 68.98 & 43.33 & 53.22 & 30.00 & 53.22 & 26.67 \\
                & AIME 2025       & 55.34 & 10.00 & 39.29 &  0.00 & 27.72 &  0.00 \\
                & Minerva-Math    & 34.13 &  2.94 & 32.19 &  0.37 & 28.18 &  0.00 \\
\midrule
                & MATH-500        & 70.98 & 47.20 & 63.69 & 29.60 & 53.84 & 15.20 \\
                & AMC-23          & 78.23 & 57.50 & 69.09 & 52.50 & 69.73 & 47.50 \\
Qwen3-8B-Base   & AIME 2024       & 77.28 & 60.00 & 55.76 & 33.33 & 57.85 & 30.00 \\
                & AIME 2025       & 51.97 & 10.00 & 38.60 &  0.00 & 32.21 &  0.00 \\
                & Minerva-Math    & 35.87 &  2.57 & 33.41 &  0.00 & 28.52 &  0.00 \\
\midrule
                & MATH-500        & 74.93 & 55.80 & 70.08 & 39.80 & 60.23 & 23.80 \\
                & AMC-23          & 75.32 & 55.00 & 73.20 & 60.00 & 75.52 & 52.50 \\
Qwen3-14B-Base  & AIME 2024       & 72.31 & 50.00 & 61.48 & 40.00 & 60.54 & 40.00 \\
                & AIME 2025       & 56.07 & 10.00 & 36.41 &  0.00 & 32.46 &  0.00 \\
                & Minerva-Math    & 38.12 &  4.41 & 34.11 &  0.74 & 29.72 &  0.00 \\
\midrule
                & MATH-500        & 80.12 & 62.40 & 75.03 & 45.60 & 64.93 & 33.00 \\
                & AMC-23          & 80.40 & 60.00 & 73.24 & 55.00 & 77.21 & 55.00 \\
Qwen3-30B-A3B   & AIME 2024       & 74.26 & 53.33 & 66.53 & 40.00 & 62.90 & 33.33 \\
                & AIME 2025       & 51.57 &  6.67 & 43.18 &  0.00 & 30.55 &  0.00 \\
                & Minerva-Math    & 37.01 &  3.68 & 34.96 &  1.10 & 29.22 &  0.00 \\
\bottomrule
\end{tabular}}%
\end{table*}

\begin{table*}[t]
\centering
\scriptsize
\caption{RougeL and exact-match (EM) scores for \textsc{Qwen2.5} and \textsc{Llama-3} on five mathematics benchmarks when each model receives only the first $x\%$ of the question ($x=80,60,40$) and must greedily complete the remainder. \textbf{Takeaway:} \textsc{Qwen2.5} shows strong reconstruction on legacy math; \textsc{Llama-3} remains low.}
\label{tab:contam_qwen25_rougel}
\resizebox{\textwidth}{!}{%
\begin{tabular}{l|l|rrrrrr}
\toprule
\textbf{Model} & \textbf{Dataset} &
\multicolumn{1}{c}{\textbf{R@80}} & \multicolumn{1}{c}{\textbf{EM@80}} &
\multicolumn{1}{c}{\textbf{R@60}} & \multicolumn{1}{c}{\textbf{EM@60}} &
\multicolumn{1}{c}{\textbf{R@40}} & \multicolumn{1}{c}{\textbf{EM@40}} \\
\midrule
                & MATH-500        & 79.63 & 61.60 & 70.30 & 41.20 & 60.81 & 25.40 \\
                & AMC-23          & 77.35 & 57.50 & 71.47 & 50.00 & 66.29 & 42.50 \\
Qwen2.5-Math-7B & AIME 2024       & 69.35 & 53.33 & 55.90 & 30.00 & 56.53 & 30.00 \\
                & AIME 2025       & 48.20 & 10.00 & 36.31 &  0.00 & 27.23 &  0.00 \\
                & Minerva-Math    & 33.42 &  2.94 & 30.41 &  0.37 & 26.88 &  0.00 \\
\midrule
                & MATH-500        & 66.09 & 38.20 & 58.15 & 18.00 & 48.34 &  6.00 \\
                & AMC-23          & 69.47 & 47.50 & 67.70 & 45.00 & 61.08 & 35.00 \\
Qwen2.5-7B      & AIME 2024       & 61.28 & 30.00 & 46.19 & 16.67 & 46.14 & 16.67 \\
                & AIME 2025       & 54.62 & 10.00 & 42.41 &  0.00 & 30.56 &  0.00 \\
                & Minerva-Math    & 33.88 &  2.94 & 31.13 &  0.00 & 27.06 &  0.00 \\
\midrule
                 & MATH-500       & 60.05 & 28.60 & 51.10 & 10.40 & 43.41 &  2.60 \\
                 & AMC-23         & 61.38 & 30.00 & 51.22 & 17.50 & 42.90 & 10.00 \\
Qwen2.5-7B-Instr & AIME 2024      & 55.41 & 20.00 & 36.28 &  0.00 & 30.64 &  0.00 \\
                 & AIME 2025      & 58.12 & 10.00 & 37.96 &  0.00 & 29.81 &  0.00 \\
                 & Minerva-Math   & 34.37 &  2.94 & 28.50 &  0.00 & 25.99 &  0.00 \\
\midrule
                & MATH-500        & 69.50 & 45.00 & 61.20 & 27.00 & 51.24 & 12.80 \\
                & AMC-23          & 74.58 & 47.50 & 68.20 & 42.50 & 67.82 & 37.50 \\
Qwen2.5-14B     & AIME 2024       & 65.42 & 40.00 & 49.76 & 23.33 & 48.97 & 23.33 \\
                & AIME 2025       & 52.56 &  6.67 & 38.02 &  0.00 & 32.92 &  0.00 \\
                & Minerva-Math    & 35.64 &  3.68 & 31.30 &  0.00 & 27.93 &  0.00 \\
\midrule
                & MATH-500        & 76.98 & 56.60 & 71.14 & 41.20 & 58.32 & 25.20 \\
                & AMC-23          & 79.18 & 62.50 & 74.59 & 60.00 & 75.78 & 55.00 \\
Qwen2.5-32B     & AIME 2024       & 73.25 & 56.67 & 60.92 & 40.00 & 60.47 & 33.33 \\
                & AIME 2025       & 53.67 &  6.67 & 40.91 &  0.00 & 31.90 &  0.00 \\
                & Minerva-Math    & 34.88 &  2.94 & 32.92 &  0.00 & 28.32 &  0.00 \\
\midrule
                & MATH-500        & 46.49 & 14.00 & 38.75 &  3.00 & 32.19 &  0.60 \\
                & AMC-23          & 40.46 &  0.00 & 29.87 &  0.00 & 25.28 &  0.00 \\
Llama-3-1.8B    & AIME 2024       & 49.70 & 16.67 & 31.45 &  0.00 & 27.13 &  0.00 \\
                & AIME 2025       & 50.16 &  6.67 & 32.97 &  0.00 & 27.18 &  0.00 \\
                & Minerva-Math    & 35.32 &  1.47 & 29.17 &  0.00 & 26.95 &  0.00 \\
\bottomrule
\end{tabular}}%
\end{table*}

\subsection{Data Contamination Probes: SimpleQA (Control)}
\label{app:contam-simpleqa}

To test whether partial-prompt reconstruction reflects a general suffix-completion ability, we repeat the probe on \textsc{SimpleQA}.
If math-set effects were general, we would expect similarly high ROUGE-L/EM at large prefixes here.
They are not: scores cluster modestly across families.
\textbf{See Table~\ref{tab:simpleqa_contam} below for the ROUGE-L and exact-match results across prefix lengths.}

\begin{table*}[t]
\centering
\scriptsize
\caption{Rouge-L and exact-match (EM) scores on \textsc{SimpleQA} when each model receives only the first $x\%$ of the question ($x=80,60,40$) and must greedily complete the remainder. \textbf{Takeaway:} No consistent \textsc{Qwen} advantage; effects seen on legacy math do not generalize, supporting the contamination interpretation.}
\label{tab:simpleqa_contam}
\resizebox{\textwidth}{!}{%
\begin{tabular}{l|l|rrrrrr}
\toprule
\textbf{Model} & \textbf{Dataset} &
\multicolumn{1}{c}{\textbf{R@80}} & \multicolumn{1}{c}{\textbf{EM@80}} &
\multicolumn{1}{c}{\textbf{R@60}} & \multicolumn{1}{c}{\textbf{EM@60}} &
\multicolumn{1}{c}{\textbf{R@40}} & \multicolumn{1}{c}{\textbf{EM@40}} \\
\midrule
Qwen2.5-Math-7B            & SimpleQA & 29.34 & 12.47 & 16.19 &  0.69 & 13.28 & 0.00 \\
Qwen2.5-7B                 & SimpleQA & 37.61 & 19.63 & 19.88 &  1.85 & 14.64 & 0.23 \\
Qwen2.5-7B-Instruct        & SimpleQA & 36.06 & 17.78 & 19.76 &  1.62 & 15.87 & 0.46 \\
Qwen2.5-14B                & SimpleQA & 39.97 & 21.94 & 21.69 &  3.23 & 15.98 & 0.23 \\
Qwen2.5-32B                & SimpleQA & 41.06 & 23.09 & 21.87 &  3.23 & 14.78 & 0.00 \\
Llama-3-1.8B               & SimpleQA & 37.11 & 19.86 & 20.24 &  2.08 & 14.22 & 0.00 \\
Qwen3-0.6B-Base            & SimpleQA & 26.45 & 10.39 & 16.35 &  0.46 & 14.42 & 0.00 \\
Qwen3-1.7B-Base            & SimpleQA & 27.55 & 12.47 & 16.50 &  0.92 & 14.97 & 0.23 \\
Qwen3-4B-Base              & SimpleQA & 33.52 & 16.40 & 17.87 &  1.85 & 14.93 & 0.00 \\
Qwen3-8B-Base              & SimpleQA & 33.99 & 16.40 & 19.23 &  1.39 & 15.39 & 0.00 \\
Qwen3-14B-Base             & SimpleQA & 37.24 & 19.40 & 21.44 &  2.31 & 16.65 & 0.23 \\
Qwen3-30B-A3B-Base         & SimpleQA & 38.80 & 19.63 & 22.07 &  3.00 & 16.55 & 0.00 \\
Qwen3-23B-A22B-Instr-2507  & SimpleQA & 53.88 & 38.34 & 34.58 & 10.39 & 22.74 & 1.39 \\
\bottomrule
\end{tabular}}%
\end{table*}

\end{document}